\documentclass[runningheads]{llncs}
\usepackage{cite}
\usepackage{graphicx}
\usepackage{threeparttable}
\usepackage{multirow}
\usepackage{booktabs}
\begin{document}
\title{A General Benchmark Framework is Dynamic Graph Neural Network Need}
%
%
\author{Yusen Zhang}
\authorrunning{Yusen Zhang}
%
\institute{University of Shanghai for Science and Technology}
\maketitle              
\begin{abstract}
Dynamic graph learning is crucial for modeling real-world systems with evolving relationships and temporal dynamics. However, the lack of a unified benchmark framework in current research has led to inaccurate evaluations of dynamic graph models. This paper highlights the significance of dynamic graph learning and its applications in various domains. It emphasizes the need for a standardized benchmark framework that captures temporal dynamics, evolving graph structures, and downstream task requirements. Establishing a unified benchmark will help researchers understand the strengths and limitations of existing models, foster innovation, and advance dynamic graph learning. In conclusion, this paper identifies the lack of a standardized benchmark framework as a current limitation in dynamic graph learning research . Such a framework will facilitate accurate model evaluation, drive advancements in dynamic graph learning techniques, and enable the development of more effective models for real-world applications.

\keywords{Graph neural network  \and Dynamic Graph \and Benchmark Framework.}
\end{abstract}
\section{Introduction}

Dynamic neural network, also called dynamic graph learning, has emerged as a prominent field in machine learning, focusing on the analysis and modeling of data that evolves over time, such as social networks, transportation systems, and financial markets \cite{feng2023towards, perozzi2014deepwalk, rossi2020temporal}. Unlike traditional static graphs, where relationships between entities remain fixed, dynamic graphs capture the temporal dynamics and evolving interactions among the elements of the graph. This temporal aspect poses unique challenges for modeling and analysis, requiring specialized techniques to capture the changing nature of the underlying data \cite{luo2022neighborhood, kipf2016variational, wang2021inductive}.

Dynamic graph learning finds extensive applications in various domains and has witnessed rapid development in recent years, such as social network analysis, traffic prediction, financial market analysis, and recommendation systems \cite{pareja2020evolvegcn, hamilton2017inductive}. The field has made significant progress in developing models, benchmark datasets, and evaluation metrics to tackle the challenges posed by evolving graph structures and temporal dynamics \cite{cui2018survey}. Ongoing research aims to further advance the scalability, efficiency, and effectiveness of dynamic graph learning models in real-world applications.

One important challenge in dynamic graph learning is the need for a unified framework to evaluate the performance of models. The dynamic nature of the data makes it crucial to assess the effectiveness of models in capturing temporal dependencies, predicting future states, and adapting to changing graph structures. However, existing evaluation techniques often fall short in addressing these specific requirements.

To address this critical gap, our core contribution is to highlight the necessity of a unified framework for evaluating dynamic graph learning models. Such a framework should encompass a range of evaluation metrics, benchmarks, and methodologies tailored to the dynamic graph context. By establishing a unified framework, researchers and practitioners can effectively compare and assess the performance of different models, enabling advancements in the field of dynamic graph learning.

In developing this unified framework, several key aspects need to be considered. First, the framework should account for the temporal dynamics of the graph, evaluating models based on their ability to capture and exploit temporal dependencies. This includes assessing the accuracy of predictions for future graph states, measuring the model's sensitivity to changes in graph structure over time, and quantifying its capability to adapt to evolving dynamics.

Second, the framework should incorporate metrics that capture the efficiency and scalability of dynamic graph learning models. As the size and complexity of dynamic graphs increase, it becomes crucial to evaluate models based on their computational requirements and the ability to handle large-scale data efficiently.

Furthermore, the unified framework should include appropriate benchmarks and datasets that reflect real-world dynamics. These benchmarks should encompass a variety of domains, allowing researchers to evaluate their models under different scenarios and settings. Additionally, the framework should consider the integration of external information sources, such as external knowledge graphs, to enhance the modeling and evaluation process.

In conclusion, dynamic graph learning presents unique challenges that demand a unified framework for model evaluation. By addressing the temporal dynamics of the data, incorporating appropriate metrics, benchmarks, and datasets, and considering efficiency and scalability, we can establish a comprehensive framework to assess the performance of dynamic graph learning models. This unified framework will foster advancements in the field, enabling the development of more accurate, adaptive, and scalable models for dynamic graph analysis and prediction.

\section{Related Work}

In this section, we provide an overview of the existing research and developments in the field of dynamic graph learning. We categorize the related work into three main areas: modeling approaches, evaluation techniques, and applications.

\subsection{Modeling Approaches}

Various modeling approaches have been proposed to tackle the challenges of dynamic graph learning \cite{zuo2018embedding}. Temporal graph neural networks (TGNs) have gained prominence for capturing temporal dependencies in dynamic graphs \cite{you2022roland, vaswani2017attention, MNCI_ML_SIGIR}. TGNs extend traditional graph neural networks by incorporating time-aware mechanisms, enabling them to model evolving graph structures and capture temporal dynamics \cite{kumar2019predicting}. Other approaches include recurrent neural networks (RNNs) and attention-based models, which focus on capturing the sequential nature of dynamic graphs \cite{xu2020inductive}.

Furthermore, research has explored techniques for graph evolution analysis to understand the changing patterns and behaviors of dynamic graphs. These techniques involve detecting important nodes and edges that undergo significant changes over time and understanding the dynamics of the graph.

\subsection{Downstream Tasks}

Evaluating the performance of dynamic graph learning models is typically done through downstream task evaluation. Researchers have used various downstream tasks to assess the effectiveness of their models \cite{grover2016node2vec, liu2023reinforcement}. These tasks include link prediction, node classification, node clustering, graph classification, graph generation, and anomaly detection, etc\cite{ hamilton2020graph, TGC_ML, S2T_ML}. By evaluating the models on these tasks, researchers can measure their ability to capture temporal dependencies, adapt to evolving graph structures, and generate accurate predictions for downstream applications.

\subsection{Applications}

Dynamic graph learning has found applications in a wide range of domains \cite{ma2022curriculum, liang2022survey, TMac_ML_MM}. In social network analysis, it has been applied to tasks such as information diffusion prediction, community detection, and recommendation systems \cite{trivedi2019dyrep, yang2023multi}. Traffic prediction and transportation systems leverage dynamic graph learning to optimize route planning and traffic flow management. Financial market analysis benefits from dynamic graph learning by predicting market trends, detecting anomalies, and optimizing investment strategies. Recommendation systems use dynamic graph models to provide personalized and timely recommendations to users.

Overall, the related work in dynamic graph learning includes a variety of modeling approaches, evaluation techniques, and applications \cite{fan2022dynamic, yu2023g}. The modeling approaches encompass techniques such as TGNs, RNNs, and attention-based models. Evaluation techniques focus on assessing the performance of models on downstream tasks, such as link prediction, node classification, and graph generation. The applications span social network analysis, traffic prediction, financial market analysis, and recommendation systems. Our work extends the existing research by emphasizing the need for comprehensive evaluation on relevant downstream tasks to demonstrate the effectiveness and applicability of dynamic graph learning models.

\section{Dynamic Graph Neural Networks}

Dynamic graph neural networks (DGNNs) provide a general framework for modeling dynamic graph data. The general model paradigm of a DGNN can be represented as follows.

At each time step t, given an input graph $G_t = (V_t, E_t, X_t)$, where $V_t$ represents the set of nodes, $E_t$ represents the set of edges, and $X_t$ denotes the node features, the DGNN updates node representations through a recurrent process. This process consists of two fundamental operations: message passing and node update.

\subsection{Message Passing}

The message passing step involves aggregating information from neighboring nodes and edges to compute new representations for each node. It can be defined as:

\begin{equation}
    m_{v,t} = AGG^t({h_{u,t-1} : u \in N(v)})
\end{equation}

Here, $h_{u,t-1}$ represents the previous hidden state of node $u$ at time $t-1$, $N(v)$ represents the neighborhood of node $v$, and $AGG^t$ is a function that aggregates the information from neighboring nodes.

\subsection{Node Update}

The node update step takes into account the aggregated messages and the current node features to compute the new hidden state of each node. It can be expressed as:

\begin{equation}
    h_{v,t} = COM^t(h_{v,t-1}, m_{v,t}, x_{v,t})
\end{equation}

Where $h_{v,t-1}$ represents the previous hidden state of node $v$ at time $t-1$, $m_{v,t}$ is the aggregated message for node $v$ at time $t$, $x_{v,t}$ denotes the node features at time $t$, and $COM^t$ is a function that combines the information.

By iteratively applying the message passing and node update steps, the DGNN captures the temporal dynamics and evolving graph structures over multiple time steps. The final hidden states of the nodes can be used for various downstream tasks, such as node classification, link prediction, or graph generation.

Overall, the general model paradigm of a DGNN involves message passing to aggregate information from neighboring nodes and edges and node update to compute new hidden states based on the aggregated messages and current node features. This framework provides a flexible and powerful approach for modeling dynamic graph data.

\section{Experiment Discussion}

The findings of this study highlight the significant variations in evaluation metrics, comparison methods, and experimental results among different dynamic graph learning algorithms, even when employing the same dataset. These variations can be attributed to several factors, including the diversity of algorithmic approaches, the inherent complexity of dynamic graph learning tasks, and the absence of standardized evaluation practices in the field.

One of the primary sources of variation lies in the selection of evaluation metrics. Different algorithms often utilize distinct metrics to assess their performance on dynamic graph learning tasks. For instance, while some algorithms may focus on accuracy-based metrics such as precision, recall, and F1-score, others may prioritize topological measures like average node degree or clustering coefficient. This discrepancy in metric selection makes it challenging to compare the performance of different algorithms objectively, as the choice of metrics can significantly impact the perceived effectiveness of a given method.

Another factor contributing to the variations is the diversity of comparison methods employed across different studies. Some algorithms may be compared against baseline methods, while others may be evaluated against state-of-the-art approaches. Additionally, the absence of a standardized benchmark dataset for dynamic graph learning further complicates the comparison process. As a result, algorithms may be evaluated using different subsets of the same dataset, making it difficult to draw conclusive comparisons between their performances.

Furthermore, the experimental results presented by different dynamic graph learning algorithms also exhibit considerable variations. Various factors, such as the specific implementation details, parameter settings, and preprocessing techniques, can influence the experimental outcomes. Moreover, the dynamic nature of graph data introduces additional complexities, as algorithms may perform differently depending on the temporal characteristics and evolution patterns present in the dataset. These variations in experimental results further highlight the need for standardized experimental setups and result reporting in dynamic graph learning research.

For example, TGN \cite{rossi2020temporal}, NAT \cite{luo2022neighborhood}, DGNN \cite{zheng2023decoupled}, CAW \cite{wang2021inductive}, and JODIE \cite{kumar2019predicting} are all utilize the Wikipedia and Reddit datasets, but the performance are different in their experiment part, as shown in Table 1.

\begin{table}[t]
    \centering
    \caption{Different metrics on same datasets.}
    \label{datasets}
    \begin{threeparttable}
       
        \begin{tabular}{c|c c c c c}
            \toprule[2pt]
            & TGN& NAT& DGNN& CAW& JODIE \\
            \midrule[1pt]
            Wikipedia& AP (98.64)& AP (98.68)& AUC (89.81)& AUC (99.89)& MRR (74.60) \\
            Reddit& AP (98.80)& AP (99.10)& AUC (67.53)& AUC (99.98)& MRR (72.60) \\
            \bottomrule[2pt]
        \end{tabular}
    \end{threeparttable}
\end{table}

To address these challenges and promote fair comparisons in the field of dynamic graph learning, it is imperative to establish standardized evaluation practices. This could involve the development of benchmark datasets that adequately capture the diverse characteristics of dynamic graph data and the definition of a set of evaluation metrics that encompass both accuracy-based and topological measures. Additionally, guidelines for experimental setups, including parameter settings and preprocessing techniques, should be established to ensure consistency across different studies. By adopting standardized evaluation practices, researchers can facilitate more reliable and consistent comparisons between dynamic graph learning algorithms, leading to more meaningful advancements in the field.

\section{Conclusion}

In conclusion, the variations in evaluation metrics, comparison methods, and experimental results among current dynamic graph learning algorithms using the same dataset pose challenges in accurately assessing their performance. Addressing these variations through the establishment of standardized evaluation practices is crucial to enable fair and meaningful comparisons, ultimately driving the progress of dynamic graph learning research.

\bibliographystyle{splncs04}
\bibliography{mybib}

\begin{thebibliography}{10}
\providecommand{\url}[1]{\texttt{#1}}
\providecommand{\urlprefix}{URL }
\providecommand{\doi}[1]{https://doi.org/#1}

\bibitem{cui2018survey}
Cui, P., Wang, X., Pei, J., Zhu, W.: A survey on network embedding. TKDE  (2018)

\bibitem{fan2022dynamic}
Fan, W., Liu, M., Liu, Y.: A dynamic heterogeneous graph perception network with time-based mini-batch for information diffusion prediction. In: International Conference on Database Systems for Advanced Applications. pp. 604--612. Springer (2022)

\bibitem{feng2023towards}
Feng, K., Li, C., Zhang, X., Zhou, J.: Towards open temporal graph neural networks. arXiv preprint arXiv:2303.15015  (2023)

\bibitem{grover2016node2vec}
Grover, A., Leskovec, J.: node2vec: Scalable feature learning for networks. In: SIGKDD (2016)

\bibitem{hamilton2017inductive}
Hamilton, W., Ying, Z., Leskovec, J.: Inductive representation learning on large graphs. NeurIPS  (2017)

\bibitem{hamilton2020graph}
Hamilton, W.L.: Graph representation learning. Synthesis Lectures on Artifical Intelligence and Machine Learning  (2020)

\bibitem{kipf2016variational}
Kipf, T.N., Welling, M.: Variational graph auto-encoders. In: NeurIPS (2016)

\bibitem{kumar2019predicting}
Kumar, S., Zhang, X., Leskovec, J.: Predicting dynamic embedding trajectory in temporal interaction networks. In: Proceedings of the 25th ACM SIGKDD international conference on knowledge discovery \& data mining. pp. 1269--1278 (2019)

\bibitem{liang2022survey}
Liang, K., Meng, L., Liu, M., Liu, Y., Tu, W., Wang, S., Zhou, S., Liu, X., Sun, F.: A survey of knowledge graph reasoning on graph types: Static, dynamic, and multimodal. arXiv preprint  (2022)

\bibitem{TMac_ML_MM}
Liu, M., Liang, K., Hu, D., Yu, H., Liu, Y., Meng, L., Tu, W., Zhou, S., Liu, X.: Tmac: Temporal multi-modal graph learning for acoustic event classification. In: Proceedings of the 31st ACM International Conference on Multimedia. pp. 3365--3374 (2023)

\bibitem{S2T_ML}
Liu, M., Liang, K., Xiao, B., Zhou, S., Tu, W., Liu, Y., Yang, X., Liu, X.: Self-supervised temporal graph learning with temporal and structural intensity alignment. arXiv preprint arXiv:2302.07491  (2023)

\bibitem{MNCI_ML_SIGIR}
Liu, M., Liu, Y.: Inductive representation learning in temporal networks via mining neighborhood and community influences. In: Proceedings of the 44th International ACM SIGIR Conference on Research and Development in Information Retrieval. pp. 2202--2206 (2021)

\bibitem{TGC_ML}
Liu, M., Liu, Y., Liang, K., Wang, S., Zhou, S., Liu, X.: Deep temporal graph clustering. arXiv preprint arXiv:2305.10738  (2023)

\bibitem{liu2023reinforcement}
Liu, Y., Liang, K., Xia, J., Yang, X., Zhou, S., Liu, M., Liu, X., Li, S.Z.: Reinforcement graph clustering with unknown cluster number. In: Proceedings of the 31st ACM International Conference on Multimedia. pp. 3528--3537 (2023)

\bibitem{luo2022neighborhood}
Luo, Y., Li, P.: Neighborhood-aware scalable temporal network representation learning. In: Learning on Graphs Conference. pp.~1--1. PMLR (2022)

\bibitem{ma2022curriculum}
Ma, J., Liu, Y., Liu, M., Han, M.: Curriculum contrastive learning for fake news detection. In: Proceedings of the 31st ACM International Conference on Information \& Knowledge Management. pp. 4309--4313 (2022)

\bibitem{pareja2020evolvegcn}
Pareja, A., Domeniconi, G., Chen, J., Ma, T., Suzumura, T., Kanezashi, H., Kaler, T., Schardl, T., Leiserson, C.: Evolvegcn: Evolving graph convolutional networks for dynamic graphs. In: Proceedings of the AAAI conference on artificial intelligence. pp. 5363--5370 (2020)

\bibitem{perozzi2014deepwalk}
Perozzi, B., Al-Rfou, R., Skiena, S.: Deepwalk: Online learning of social representations. In: SIGKDD (2014)

\bibitem{rossi2020temporal}
Rossi, E., Chamberlain, B., Frasca, F., Eynard, D., Monti, F., Bronstein, M.: Temporal graph networks for deep learning on dynamic graphs. arXiv preprint arXiv:2006.10637  (2020)

\bibitem{trivedi2019dyrep}
Trivedi, R., Farajtabar, M., Biswal, P., Zha, H.: Dyrep: Learning representations over dynamic graphs. In: International conference on learning representations (2019)

\bibitem{vaswani2017attention}
Vaswani, A., Shazeer, N., Parmar, N., Uszkoreit, J., Jones, L., Gomez, A.N., Kaiser, L., Polosukhin, I.: Attention is all you need. NeurIPS  (2017)

\bibitem{wang2021inductive}
Wang, Y., Chang, Y.Y., Liu, Y., Leskovec, J., Li, P.: Inductive representation learning in temporal networks via causal anonymous walks. arXiv preprint arXiv:2101.05974  (2021)

\bibitem{xu2020inductive}
Xu, D., Ruan, C., Korpeoglu, E., Kumar, S., Achan, K.: Inductive representation learning on temporal graphs. arXiv preprint arXiv:2002.07962  (2020)

\bibitem{yang2023multi}
Yang, P., Ma, J., Liu, Y., Liu, M.: Multi-modal transformer for fake news detection. Mathematical Biosciences and Engineering: MBE  \textbf{20}(8),  14699--14717 (2023)

\bibitem{you2022roland}
You, J., Du, T., Leskovec, J.: Roland: graph learning framework for dynamic graphs. In: Proceedings of the 28th ACM SIGKDD Conference on Knowledge Discovery and Data Mining. pp. 2358--2366 (2022)

\bibitem{yu2023g}
Yu, H., Ma, C., Liu, M., Liu, X., Liu, Z., Ding, M.: Guardfl: Safeguarding federated learning against backdoor attacks through attributed client graph clustering. arXiv preprint arXiv:2306.04984  (2023)

\bibitem{zheng2023decoupled}
Zheng, Y., Wei, Z., Liu, J.: Decoupled graph neural networks for large dynamic graphs. arXiv preprint arXiv:2305.08273  (2023)

\bibitem{zuo2018embedding}
Zuo, Y., Liu, G., Lin, H., Guo, J., Hu, X., Wu, J.: Embedding temporal network via neighborhood formation. In: Proceedings of the 24th ACM SIGKDD international conference on knowledge discovery \& data mining. pp. 2857--2866 (2018)

\end{thebibliography}
\end{document}